%% file: main.tex
\documentclass[11pt,journal,cspaper,compsoc,twoside]{IEEEtran}

\usepackage{color,xspace}
\usepackage{nicefrac}
\usepackage{graphicx}
\usepackage{booktabs,colortbl}%,xcolor}
\usepackage{hyperref}
\usepackage{amsmath}
\usepackage{amssymb}
\usepackage{multirow}
\usepackage{textgreek}
\usepackage{wrapfig}
\usepackage{etoolbox}

\newtoggle{tech_report}
\newtoggle{review_copy}
\newtoggle{tight}

\toggletrue{tech_report}
\nottoggle{tech_report}{%
\input{bmvc_preamble.tex}
}{%
\newcommand{\bmvaOneDot}{.\xspace}
}

\newcommand{\figref}[1]{Fig.~\ref{#1}}
\newcommand{\tblref}[1]{Table~\ref{#1}}
\newcommand{\sref}[1]{Sect.~\ref{#1}}

\renewcommand{\rowcolor}[1]{} %disabled as it breaks my latex
\definecolor{tableShade}{rgb}{0.98,0.98,0.98}

\def\eg{\emph{e.g}\bmvaOneDot}

\def\etal{\emph{et al}\bmvaOneDot}
\def\ie{\emph{i.e}\bmvaOneDot}
\def\vs{\emph{vs}\bmvaOneDot}

\newcommand{\bx}{\mathbf{x}}
\newcommand{\bu}{\mathbf{u}}
\newcommand{\bv}{\mathbf{v}}
\newcommand{\bw}{\mathbf{w}}

\newcommand{\enc}{\phi}

\newcommand{\enccnn}{\enc_{\mathrm{CNN}}}

\newcommand{\encfisher}{\enc_{\mathrm{FV}}}

\newcommand{\real}{\mathrm{R}}

\newcommand{\image}{I}

\iftoggle{tight}{%
\input{compactify.tex}
}{}

\begin{document}

\nottoggle{tech_report}{%

\maketitle

\begin{abstract}
\input{abstract.tex}
\end{abstract}

}{%

\input{ieeetran_preamble.tex}
\IEEEcompsoctitleabstractindextext{%
\begin{abstract}
\input{abstract.tex}
\end{abstract}}

\maketitle
\IEEEdisplaynotcompsoctitleabstractindextext

}

%------------------------------------------------------------------------- 

% The interest in the application of convolutional neural networks (CNNs) to image
% classification and retrieval has increased greatly over the past two years,
% following their proven good performance on large-scale classification
% benchmarks, such as the ImageNet ILSVRC dataset~\cite{Krizhevsky12}. As a
% result, a number of CNN-based methods have been
% proposed~\cite{Zeiler13,Donahue13,Sermanet14,Oquab14}, which apply ImageNet CNN
% features to smaller datasets, such as PASCAL VOC~\cite{Everingham10}.
% It is often unclear how these methods compare with each other, or previous
% state-of-the-art image classification approaches, based on shallow feature
% encodings~\cite{Perronnin10a,Chatfield11}.

%------------------------------------------------------------------------- 

\section{Introduction}\label{sec:intro}

\nottoggle{tech_report}{%
Perhaps
}{%
\IEEEPARstart{P}{erhaps}
}%
the single most important design choice in current
state-of-the-art image classification and object recognition systems
is the choice of visual features, or image representation. In fact,
most of the quantitative improvements to image understanding obtained
in the past dozen years can be ascribed to the introduction of
improved representations, from the \emph{Bag-of-Visual-Words}
(BoVW)~\cite{Csurka04,Sivic03} to the \emph{(Improved) Fisher Vector}
(IFV)~\cite{Perronnin10a}. A common characteristic of these methods is
that they are largely \emph{handcrafted}. They are also relatively
simple, comprising dense sampling of local image patches, describing
them by means of visual descriptors such as SIFT, encoding them into a
high-dimensional representation, and then pooling over the image.
Recently, these handcrafted approaches have been substantially
outperformed by the introduction of the latest
generation of \emph{Convolutional Neural Networks}
  (CNNs)~\cite{Lecun89} to the computer vision field.
These networks have a substantially more
sophisticated structure than standard representations, comprising
several layers of non-linear feature extractors, and are therefore
said to be \emph{deep} (in contrast, classical representation will be
referred to as \emph{shallow}). Furthermore, while their structure is
handcrafted, they contain a very large number of parameters learnt
from data. When applied to standard image classification and object detection benchmark
datasets such as ImageNet ILSVRC~\cite{Deng09} and PASCAL
VOC~\cite{Everingham10} such networks have demonstrated excellent
performance~\cite{Donahue13,Oquab14,Razavian14,Sermanet14,Girshick14},
significantly better than standard image encodings~\cite{Chatfield11}.

%%For example, the CNN of~\cite{Krizhevsky12} substantially
%outperformed IFV in the ILSVRC 2012 image classification
%challenge. Likewise, the classification accuracy of image
%representations on PASCAL VOC was 53.4\% mAP for the BoVW model in
%2008, 61.7\% for the IFV in 2010~\cite{Chatfield11}, and 72.7\%/73.4\%
%\rk{$\leftarrow$ TODO: choose augmented or non-augmented figure} for
%DeCAF~\cite{Donahue13} and similar~\cite{Oquab14,Razavian14} CNN-based
%method introduced in late 2013. CNNs have also demonstrated
%exceptional performance in object category detection, both in
%ILSVRC~\cite{Sermanet14} and PASCAL~\cite{Girshick14}.

Despite these impressive results, it remains unclear how different
deep architectures compare to each other and to shallow computer
vision methods such as IFV. Most papers did not test these
representations extensively on a common ground, so a systematic
evaluation of the effect of different design and implementation
choices remains largely missing.  As noted in our previous work~\cite{Chatfield11},
which compared the performance of various shallow
visual encodings, the \emph{performance of computer vision systems
  depends significantly on implementation details}. For example,
state-of-the-art methods such as~\cite{Krizhevsky12} not only involve
the use of a CNN, but also include other improvements such as the use
of very large scale datasets, GPU computation, and data augmentation
(also known as data jittering or virtual sampling). These improvements
could also transfer to shallow representations such as the IFV,
potentially explaining a part of the performance gap~\cite{Paulin14}. 
%Another point of
%contention is that, when applied to small datasets such as PASCAL VOC,
%CNNs require pre-training on large external datasets to perform well.

In this study we analyse and  empirically clarify these issues,
conducting a large set of rigorous experiments (Sect.~\ref{sec:analysis}), in many ways picking
up the story where it last ended in~\cite{Chatfield11} with the
comparison of shallow encoders. We focus on methods to construct
\emph{image representations}, \ie encoding functions $\enc$ mapping an
image $\image$ to a vector $\enc(\image)\in\mathbb{R}^d$
suitable for analysis with a linear classifier, such as an SVM.  We
consider {\bf three scenarios} (Sect.~\ref{sec:methods}, Sect.~\ref{sec:details}): shallow image representations, deep representations pre-trained on outside data, and deep representation pre-trained and then fine-tuned on the target dataset.
%  In the
%first scenario, we explore a state-of-the-art shallow image
%representation, namely the IFV~\cite{Chatfield11}.  In the second
%scenario, we explore deep image features, computed using CNNs.  As in
%~\cite{Donahue13,Zeiler13}, a CNN is pre-trained as a classifier on a
%large image dataset such as ILSVRC.  After pre-training, the last
%layer of the CNN is removed and the output of the remaining
%architecture is used as a visual feature $\enc(\image)$. The feature
%is then used to perform classification in another, unrelated dataset
%such as PASCAL VOC.  The third and last scenario is similar to the
%second, except that the whole deep network is fine-tuned on the target
%dataset.
%
%In addition to testing these fundamental scenarios, we make several
%additional contributions. First, we systematically evaluate
As part of our tests, we explore {\bf
  generally-applicable best practices} that are nevertheless more
often found in combination with CNNs~\cite{Krizhevsky12} or,
alternatively, with shallow encoders~\cite{Chatfield11}, porting them
with mutual benefit. These are (Sect.~\ref{sec:methods}): the use of
\emph{colour information}, feature \emph{normalisation}, and, most
importantly, the use of \emph{substantial data augmentation}. We also
determine {\bf scenario-specific best-practices},
improving the ones in~\cite{Chatfield11,Perronnin12} and others, including dimensionality
reduction for deep features. 
%For
%example, we evaluate the effect of spatially extending SIFT
%descriptors in IFV~\cite{Sanchez12} and the effect of choosing
%different network structures, and in particular network sizes, in
%CNN.
Finally, we achieve {\bf performance competitive with the state of the art~\cite{Oquab14a,Wei14}} on
PASCAL VOC classification using less additional training data and significantly simpler techniques.
As in~\cite{Chatfield11},
the source code and models to reproduce all experiments in this paper is available on
the project website\footnote{\url{http://www.robots.ox.ac.uk/~vgg/research/deep_eval/}}.

\iftoggle{tight}{\vspace{-0.4em}}{}
%------------------------------------------------------------------------- 
\section{Scenarios}\label{sec:methods}
%------------------------------------------------------------------------- 
\iftoggle{tight}{\vspace{-0.3em}}{}

This section introduces the three types of image representation $\enc(\image)$
considered in this paper, describing them within the context of three different
scenarios. Having outlined details specific to each, general methodologies which
apply to all three scenarios are reviewed, such as data augmentation and feature
normalisation, together with the linear classifier (trained with a standard
hinge loss). We also specify here the benchmark datasets used in the evaluation.

\iftoggle{tight}{\vspace{-0.2em}}{}
\subsectionpar{Scenario 1: Shallow representation (IFV)} Our reference
shallow image representation is the IFV~\cite{Perronnin10a}. Our
choice is motivated by the fact that IFV usually outperforms related
encoding methods such as BoVW, LLC~\cite{Chatfield11}, and
VLAD~\cite{Jegou12a}. Given an image $\image$, the IFV
$\encfisher(\image)$ is obtained by extracting a dense collection of
patches and corresponding local descriptors $\bx_i\in\real^{D}$ (\eg SIFT) from the image at multiple scales. 
Each descriptor $\bx_i$ is
then soft-quantized using a Gaussian Mixture Model with $K$
components. First and second order differences between each descriptor
$\mathbf{x}_i$ and its Gaussian cluster mean $\mu_k$ are accumulated
in corresponding blocks $\bu_k$, $\bv_k$ in the vector
$\encfisher(\image)\in\real^{2KD}$, appropriately weighed by the
Gaussian soft-assignments and covariance, leading to a
$2KD$-dimensional image representation
$
\encfisher(\image) = [\bu_1^\top, \bv_1^\top, \dots \bu_K^\top, \bv_K^\top ]^\top.
$
The \emph{improved} version of the Fisher vector involves
post-processing $\encfisher$ by computing the signed square-root of
its scalar components and normalising the result to a unit $\ell^2$
norm. The details of this construction can be found
in~\cite{Perronnin10a}; here we follow the notation
of~\cite{Chatfield11}.

\iftoggle{tight}{\vspace{-0.2em}}{}
\subsectionpar{Scenario 2: Deep representation (CNN) with pre-training}
Our deep representations are inspired by the success of the CNN of
Krizhevsky~\etal~\cite{Krizhevsky12}. As shown
in~\cite{Donahue13,Zeiler13}, the vector of activities
$\enccnn(\image)$ of the penultimate layer of a deep CNN, learnt on a
large dataset such as ImageNet~\cite{Deng09}, can be used as a
powerful image descriptor applicable to other datasets. Numerous
CNN architectures that improve the previous state of the art obtained
using shallow representations have been proposed, but choosing the best
one remains an open question. Many are inspired by~\cite{Krizhevsky12}:
DeCAF~\cite{Donahue13,Girshick14}, Caffe~\cite{Jia13},
Oquab~\etal~\cite{Oquab14}. Others use larger networks with a smaller stride of the first convolutional layer: Zeiler and Fergus~\cite{Zeiler13} and
OverFeat~\cite{Sermanet14,Razavian14}. Other differences include
the CNN pre-training protocols.
% :~\cite{Donahue13,Oquab14} 
% did not use the colour space jittering
% of~\cite{Krizhevsky12}, and~\cite{Donahue13} also resized all input
% images to the same size, distorting their aspect ratio.  
% It is also
% worth noting that some publicly available CNN
% implementations~\cite{Donahue13,Sermanet14} do not contain the
% training code, which prevents their usage in the scenarios where CNN
% fine-tuning (re-training initialised by a pre-trained network) is
% required. 
Here we adopt a single learning framework and experiment
with architectures of different complexity exploring their performance-speed trade-off.

\iftoggle{tight}{\vspace{-0.2em}}{}
\subsectionpar{Scenario 3: Deep representation (CNN) with pre-training and fine-tuning}
In Scenario~2 features are trained on one (large) dataset and applied to
another (usually smaller). However, it was demonstrated~\cite{Girshick14} that fine-tuning a pre-trained CNN
on the target data can significantly improve the performance. We consider this 
scenario separately from that of Scenario~2, as the image features become dataset-specific after the fine-tuning.

\iftoggle{tech_report}{%
\subsection{Commonalities}
\noindent We now turn to what is in common across the scenarios.
}

% \rk{KS: add a transition sentence here to separate the scenarios}
\iftoggle{tight}{\vspace{-0.2em}}{}
\subsubsectionpar{Data augmentation} Data augmentation is  a method applicable to shallow
and deep representations, but that has been so far mostly applied to
the latter~\cite{Krizhevsky12,Zeiler13}. By augmentation we mean perturbing an image $\image$
by transformations that leave the underlying class unchanged (\eg
cropping and flipping) in order to generate additional examples of the
class.  Augmentation can be applied at training time, at test time, or
both. The augmented samples can either be taken as-is %for use during training
or %can be
combined to form a single feature, e.g.\ using 
sum/max-pooling or stacking.

\iftoggle{tight}{\vspace{-0.2em}}{}
\subsubsectionpar{Linear predictors}
All the representations $\enc(\image)$ in the three scenarios are used
to construct \emph{linear predictors} $\langle \bw,
\enc(\image)\rangle$ for each class to be recognized. These predictors
are learnt using Support Vector Machines (SVM) by fitting $\bw$
to the available training data by minimizing an objective function
balancing a quadratic regularizer and the hinge-loss. The parameter $C$ in the SVM,
trading-off regularizer and loss, is determined using an held-off validation
subset of the data. Here we use the same learning framework with all representations.
It is common
experience that linear classifiers are particularly sensitive to the
\emph{normalisation of the data} and that, in particular, SVMs tend to
benefit from $\ell^2$ normalisation~\cite{Perronnin10a} (an interpretation is that after normalisation the inner product corresponds to the cosine similarly). 
%Here we
%investigate whether normalisation is important for CNN-based features
%as well.

\iftoggle{tight}{\vspace{-0.2em}}{}
\subsectionpar{Benchmark data} As reference benchmark we use the PASCAL
VOC~\cite{Everingham10} data as already done in~\cite{Chatfield11}. The \textbf{VOC-2007} edition contains about
10,000 images split into train, validation, and test sets, and
labelled with twenty object classes. A one-vs-rest SVM classifier for
each class is learnt and evaluated independently and the performance
is measured as mean Average Precision (mAP) across all classes. The
\textbf{VOC-2012} edition contains roughly twice as many images and does not
include test labels; instead, evaluation uses the official PASCAL
Evaluation Server. To
train deep representations we use
the {\bf ILSVRC-2012} challenge dataset. This contains 1,000
object categories from ImageNet~\cite{Deng09} with roughly 1.2M
training images, 50,000 validation images, and 100,000 test
images. Performance is evaluated using the top-5 classification
error. % in the sense that a prediction is considered correct if the ground-truth image label is contained within the top-5 predicted categories.
Finally, we also evaluate over the \textbf{Caltech-101} and \textbf{Caltech-256} image classification benchmarks~\cite{FeiFei04,Griffin07}.
For Caltech-101, we followed the protocol of~\cite{Chatfield11}, and considered three random splits into training and testing data, each 
of which comprises 30 training and up to 30 testing images per class. For Caltech-256, two random splits were generated, each of which contains
60 training images per class, and the rest are used for testing. On both Caltech datasets, performance is measured using mean class accuracy.

% All the experiments use a linear SVM on top of each encoding. The
% parameter $C$ of the SVM (regularization-loss trade off) is
% determined using the provided validation split of the PASCAL VOC
% data. ***KAREN - add something on CNN classif layers ***

\iftoggle{tight}{\vspace{-0.5em}}{}
%------------------------------------------------------------------------- 
\section{Details}\label{sec:details}
%-------------------------------------------------------------------------
\iftoggle{tight}{\vspace{0.8em}}{}

\iftoggle{tech_report}{%
This section gives the implementation details of the methods
introduced in Sect.~\ref{sec:methods}.
}

%------------------------------------------------------------------------- 
\subsection{Improved Fisher Vector details}\label{sec:ifv}
%------------------------------------------------------------------------- 

Our IFV representation uses a slightly improved setting compared
to the best result of~\cite{Chatfield11}. \goodbreak Computation starts by upscaling the image
$\image$ by a factor of 2~\cite{Sanchez12}, followed by SIFT features extraction with a
stride of 3 pixels at 7 different scales with $\sqrt{2}$ scale
increments. These features are square-rooted as suggested
by~\cite{Arandjelovic12}, and decorrelated and reduced in dimension
from $128D$ to $80D$ using PCA. A GMM with $K=256$ components is
learnt from features sampled from the training images.  Hence the
Fisher Vector $\encfisher(I)$ has dimension $2KD = 40,960$. Before use
in classification, the vector is signed-square-rooted and
$l^2$-normalised (square rooting correspond to the Hellinger's kernel
map~\cite{Vedaldi11}). As in~\cite{Chatfield11}, square-rooting is
applied twice, once to the raw encodings, and once again after sum
pooling and normalisation.
%IFV is an \emph{orderless} representation, in the sense that it does
%not capture the location of the measured features in the image, and
%hence retains almost no geometric information. In order to capture at
%least a weak geometry, the IFV representation is used in a
%\emph{spatial pyramid}
%scheme~\cite{Lazebnik06}. As in~\cite{Chatfield11}, the image is
%divided into $1\times 1$, $3 \times 1$, and $2\times 2$ spatial
%subdivisions and corresponding IFV are computed and stacked for an
%overall dimension of $8\times 2KD=327,680$ elements.
In order to capture weak geometrical information, the IFV representation is used in a
\emph{spatial pyramid}~\cite{Lazebnik06}. As in~\cite{Chatfield11},
the image is divided into $1\times 1$, $3 \times 1$, and $2\times 2$
spatial subdivisions and corresponding IFVs are computed and stacked with an
overall dimension of $8\times 2KD=327,680$ elements.

In addition to this standard formulation, we experiment with a few
modifications. The first one is the use of \emph{intra-normalisation}
of the descriptor blocks, an idea recently proposed for the VLAD
descriptor~\cite{arandjelovic13}.  In this case, the $\ell^2$
normalisation is applied to the individual sub-blocks $(\bu_k,\bv_k)$
of the vector $\encfisher(\image)$, which helps to alleviate the local
feature burstiness~\cite{Jegou09a}. In the case of the improved
intra-normalised features, it was found that applying the square-rooting
only once to the final encoding produced the best results.

The second modification is the use of \emph{spatially-extended local
descriptors}~\cite{Sanchez12} instead of a spatial pyramid. Here
descriptors $\bx_i$ are appended with their image location $(x_i,y_i)$
before quantization with the GMM. Formally, $\bx_i$ is extended, after
PCA projection, with its normalised spatial coordinates: $[\bx_i^\top,
x_i/W-0.5, y_i/H-0.5]^\top$, where $W \times H$ are the dimensions of
the image.  Since the GMM quantizes both appearance and location, this
allows for spatial information to be captured directly by the
soft-quantization process. This method is significantly more
memory-efficient than using a spatial pyramid. Specifically, the
PCA-reduced SIFT features are spatially augmented by appending $(x,y)$
yielding $D=82$ dimensional descriptors pooled in a $2KD = 41,984$ dimensional
IFV.

The third modification is the use of colour features in addition to
SIFT descriptors. While colour information is used in
CNNs~\cite{Krizhevsky12} and by the original FV
paper~\cite{Perronnin10a}, it was not explored in our previous
comparison~\cite{Chatfield11}. We do so here by adopting the same
Local Colour Statistics (LCS) features as used by~\cite{Perronnin10a}.
LCS is computed by dividing an input patch into a $4\times 4$ spatial
grid (akin to SIFT), and computing the mean and variance of each of
the \emph{Lab} colour channels for each cell of the grid. The LCS
dimensionality is thus $4 \times 4 \times 2 \times 3=96$. This is then
encoded in a similar manner to SIFT.

\iftoggle{tight}{\vspace{-0.5em}}{}
%------------------------------------------------------------------------- 
\subsection{Convolutional neural networks details}\label{sec:cnn}
%------------------------------------------------------------------------- 

\begin{table}[t]
\begin{center}
 \small
 \begin{tabular}{|c|c|c|c|c|c|c|c|c|}
  \hline 
 Arch. & conv1 & conv2 & conv3 & conv4 & conv5 & full6 & full7 & full8 \\ 
  \hline \hline
%   \hline \multicolumn{8}{|c|}{\textbf{Fast CNN}} \\ \hline
  \multirow{3}{*}{CNN-F} & 64x11x11 & 256x5x5 & 256x3x3 & 256x3x3 & 256x3x3 & 4096 & 4096 & 1000 \\ 
                        & st. 4, pad 0 & st. 1, pad 2  & st. 1, pad 1 & st. 1, pad 1 & st. 1, pad 1 & drop- & drop- & soft-  \\ 
                        & LRN, x2 pool & LRN, x2 pool  & - & - & x2 pool & out & out & max  \\ 
  \hline \hline
%   \hline \multicolumn{8}{|c|}{\textbf{Medium CNN}} \\ \hline
  \multirow{3}{*}{CNN-M} & 96x7x7 & 256x5x5 & 512x3x3 & 512x3x3 & 512x3x3 & 4096 & 4096 & 1000 \\ 
                          & st. 2, pad 0 & st. 2, pad 1  & st. 1, pad 1 & st. 1, pad 1 & st. 1, pad 1 & drop- & drop- & soft- \\ 
                          & LRN, x2 pool & LRN, x2 pool  & - & - & x2 pool & out & out & max  \\ 
  \hline \hline
%   \hline \multicolumn{8}{|c|}{\textbf{Slow CNN}} \\ \hline
  \multirow{3}{*}{CNN-S} & 96x7x7 & 256x5x5 & 512x3x3 & 512x3x3 & 512x3x3 & 4096 & 4096 & 1000 \\ 
                        & st. 2, pad 0 & st. 1, pad 1 & st. 1, pad 1 & st. 1, pad 1 & st. 1, pad 1 & drop- & drop- & soft- \\ 
                        & LRN, x3 pool & x2 pool  & - & - & x3 pool & out & out & max \\ 
  \hline  
 \end{tabular}
\end{center}
\iftoggle{tight}{\vspace{-0.7em}}{}
\caption{\textbf{CNN architectures.} Each architecture contains 5 convolutional layers (conv\,1--5) and three fully-connected layers (full\,1--3).
The details of each of the convolutional layers are given in three sub-rows: the first specifies the number of convolution filters and their receptive field size as ``num x size x size'';
the second indicates the convolution stride (``st.'') and spatial padding (``pad''); the third indicates if Local Response Normalisation (LRN)~\cite{Krizhevsky12} is applied,
and the max-pooling downsampling factor.
For full\,1--3, we specify their dimensionality, which is the same for all three architectures. Full6 and full7 are regularised using dropout~\cite{Krizhevsky12},
while the last layer acts as a multi-way soft-max classifier. The activation function for all weight layers (except for full8) is the REctification Linear Unit (RELU)~\cite{Krizhevsky12}.
\iftoggle{tight}{\vspace{-0.3em}}{}
}
\label{tab:CNN_arch}
\end{table}

The CNN-based features are based on three CNN architectures representative of the state of the art (shown in \tblref{tab:CNN_arch}) each exploring a different accuracy/speed trade-off. 
To ensure a fair comparison between them, these networks are trained using the same training protocol and the same implementation, 
which we developed based on the open-source Caffe framework~\cite{Jia13}. $\ell^2$-normalising the CNN features $\enccnn(\image)$ before use in the SVM was found to be important for performance.

%\subsubsectionpar{CNN architectures }%}{}%
%The three CNN architectures (\tblref{tab:CNN_arch}) explore different accuracy/speed trade-off points.
%The trade-off is mainly controlled by choosing the filter sizes and the downsampling factors in the convolutional and pooling layers, as detailed next. The comparison with the state-of-the-art CNNs is presented in~\sref{sec:analysis}.

Our \textbf{Fast (CNN-F)} architecture is similar to the one used by Krizhevsky~\etal~\cite{Krizhevsky12}. It comprises 8 learnable layers, 5 of which are convolutional, and the last 3 are fully-connected.
The input image size is $224 \times 224$.
Fast processing is ensured by the 4 pixel stride in the first convolutional layer. The main differences between our architecture and that of~\cite{Krizhevsky12} are the reduced 
number of convolutional layers and the dense connectivity between convolutional layers (\cite{Krizhevsky12} used sparse connections to enable training on two GPUs).

Our \textbf{Medium (CNN-M)} architecture is similar to the one used by Zeiler and Fergus~\cite{Zeiler13}. It is characterised by the decreased stride and smaller receptive field
of the first convolutional layer, which was shown to be beneficial on the ILSVRC dataset.
At the same time, conv2 uses larger stride (2 instead of 1) to keep the computation time reasonable. The main difference between our net and that of~\cite{Zeiler13} is we use less filters in the conv4 layer (512 \vs~1024).

Our \textbf{Slow (CNN-S)} architecture is related to the `accurate' network from the OverFeat package~\cite{Sermanet14}. It also uses $7\times7$ filters with stride $2$ in conv1.
Unlike CNN-M and~\cite{Zeiler13}, the stride in conv2 is smaller (1 pixel), but the max-pooling window in conv1 and conv5 is larger ($3\times 3$) to compensate for the
increased spatial resolution.
Compared to~\cite{Sermanet14}, we use 5 convolutional layers as in the previous architectures (\cite{Sermanet14} used 6), and less filters in conv5 (512 instead of 1024);
we also incorporate an LRN layer after conv1 (\cite{Sermanet14} did not use contrast normalisation).

\iftoggle{tight}{\vspace{-0.2em}}{}
\subsubsectionpar{CNN training}\label{sec:CNN_train}
In general, our CNN training procedure follows that of~\cite{Krizhevsky12}, learning on ILSVRC-2012 using gradient descent with momentum.
The hyper-parameters are the same as used by~\cite{Krizhevsky12}: momentum $0.9$; weight decay $5 \cdot 10^{-4}$; initial learning rate $10^{-2}$, which is decreased by a factor of $10$, when the validation
error stop decreasing. The layers are initialised from a Gaussian distribution with a zero mean and variance equal to $10^{-2}$.
We also employ similar data augmentation in the form of random crops, horizontal flips, and RGB colour jittering.
Test time crop sampling is discussed in~\sref{sec:other}; at training time, $224 \times 224$ crops are sampled randomly, rather than deterministically.
Thus, the only notable difference to~\cite{Krizhevsky12} is that the crops are taken from the whole training image $P \times 256, P \ge 256$, rather than its $256 \times 256$ centre. 
%This results in a better coverage of images with a large aspect ratio.
%When evaluating the networks on ILSVRC-2012 (see Sect.\,4), the 10-fold augmentation of Sect.\,3.3 is employed.
Training was performed on a single NVIDIA GTX Titan GPU
% using a modified version of the publicly-available Caffe framework~\cite{Jia13}.
and the training time varied
from 5 days for CNN-F to 3 weeks for CNN-S.% (which required more iterations to converge).

\iftoggle{tight}{\vspace{-0.2em}}{}
\subsubsectionpar{CNN fine-tuning on the target dataset} 
%Given a pre-trained CNN, it can be fine-tuned on a target dataset by using it as a starting point for training.
In our experiments, we fine-tuned CNN-S using VOC-2007, VOC-2012, or Caltech-101 as the target data. 
Fine-tuning was carried out using the same framework (and the same data augmentation), as we used for CNN training on ILSVRC.
The last fully-connected layer (conv8) has output dimensionality equal to the number of classes, which differs between datasets, so we initialised it from a Gaussian distribution (as used for CNN training above).
%The last fully-connected layer (conv8) has output dimensionality equal to the number of classes (20 in the case of VOC datasets, 102 in the case of Caltech-101), so it could not 
%be initialised with the corresponding layer of the ILSVRC net; instead, we initialised it from a Gaussian distribution (as used for CNN training above).
Now we turn to dataset-specific fine-tuning details.

\noindent\textbf{VOC-2007 and VOC-2012.\;}
Considering that PASCAL VOC is a multi-label dataset (\ie~a single image might have multiple labels), we replaced the softmax regression loss
with a more appropriate loss function, for which we considered two options: one-vs-rest classification hinge loss (the same loss as used in the SVM experiments)
and ranking hinge loss. Both losses define constraints on the scores of positive ($\image_{pos}$) and negative ($\image_{neg}$) images for each class:
$w_c \phi(\image_{pos}) > 1 - \xi, w_c \phi(\image_{neg}) < -1 + \xi$ for the classification loss,
$w_c \phi(\image_{pos}) > w_c \phi(\image_{neg}) + 1 - \xi$ for the ranking loss ($w_c$ is the $c$-th row of the last fully-connected layer, which can be seen as a linear
classifier on deep features $\phi(\image)$; $\xi$ is a slack variable).
% We implemented both losses as the layers of our CNN training framework.
Our fine-tuned networks are denoted as ``CNN S TUNE-CLS'' (for the classification loss) and ``CNN S TUNE-RNK'' (for the ranking loss).
In the case of both VOC datasets, the training and validation subsets were combined to form a single training set. Given the smaller size of the training data when compared
to ILSVRC-2012, we controlled for over-fitting by using lower initial learning rates
for the fine-tuned hidden layers. The learning rate schedule for the last layer / hidden layers was:
$10^{-2}/10^{-4}\rightarrow 10^{-3}/10^{-4} \rightarrow 10^{-4}/10^{-4} \rightarrow 10^{-5}/10^{-5}$. 
%The values of other hyper-parameters were kept the same as in ILSVRC-2012 training. 

\noindent\textbf{Caltech-101 } dataset contains a single class label per image, so fine-tuning was performed using the softmax regression loss.
Other settings (including the learning rate schedule) were the same as used for the VOC fine-tuning experiments.

\iftoggle{tight}{\vspace{-0.2em}}{}
\subsubsectionpar{Low-dimensional CNN feature training}
Our baseline networks (Table~\ref{tab:CNN_arch}) have the same dimensionality of the last hidden layer (full7): 4096.
This design choice is in accordance with the state-of-the-art architectures~\cite{Krizhevsky12,Zeiler13,Sermanet14}, and leads to a 4096-D dimensional image representation,
which is already rather compact compared to IFV.
%At the same time, for large-scale image recognition applications, even lower dimensionality can be required. To investigate the dependency of the performance on the feature dimensionality,
We further trained three modifications of the CNN-M network, with lower dimensional full7 layers of: 2048, 1024, and 128 dimensions respectively.
The networks were learnt on ILSVRC-2012.
To speed-up training, all layers aside from full7/full8 were set to those of the
CNN-M net and a lower initial learning rate of $10^{-3}$ was used. The initial learning
rate of full7/full8 was set to $10^{-2}$.
%
%The performance of deep low-dimensional features is discussed in \sref{sec:analysis}.
%}{}

%\iftoggle{tight}{\vspace{-0.5em}}{}
%-------------------------------------------------------------------------
\subsection{Data augmentation details}\label{sec:other}
%-------------------------------------------------------------------------
%\iftoggle{tight}{\vspace{-0.5em}}{}

We explore three data augmentation strategies. The first
strategy is to use {\bf no augmentation}. 
%The IFV representation can take
%images of any size as input, and therefore in this case the image is processed untransformed. 
In contrast to IFV, however, CNNs require images to be transformed to a fixed
size ($224 \times 224$) even when no augmentation is used. Hence the image is downsized so that the smallest
dimension is equal to $224$ pixels and a $224 \times 224$ crop is extracted from
the centre.%of the image and used as the input to a CNN.
\footnote{Extracting a $224 \times 224$ centre crop from a $256 \times 256$ image~\cite{Krizhevsky12} resulted in worse performance.} 
The second strategy is to use {\bf flip augmentation}, mirroring 
images about
the $y$-axis producing two samples from each image. 
The third strategy, termed {\bf C+F augmentation}, combines cropping and flipping. For CNN-based representations, the image is downsized so that the
smallest dimension is equal to $256$ pixels. Then
$224 \times 224$ crops are extracted from the four corners and the centre of the image. 
Note that the crops are sampled from the whole image, rather than its $256\times 256$ centre, as done by~\cite{Krizhevsky12}.
These crops are then flipped about the $y$-axis, producing $10$ perturbed samples per input image. In the case
of the IFV encoding, the same crops are extracted, but at the original image resolution.

\iftoggle{tight}{\vspace{-0.4em}}{}
%------------------------------------------------------------------------- 
\section{Analysis}\label{sec:analysis}
%------------------------------------------------------------------------- 
\iftoggle{tight}{\vspace{-0.1em}}{}

\makeatletter
\newcommand*{\greek}[1]{%
  \expandafter\@greek\csname c@#1\endcsname
}
\newcommand*{\@greek}[1]{%
  \ifcase#1\or a\or b\or c\or d\or e\or f\or g\or h\or i\or j\or k\or l\or m\or n\or o\or p\or q\or r\or s\or t\or u\or v\or w\or x\or y\or z\or\textalpha\or\textbeta\or\textgamma\or\textdelta
    \or\textzeta\or\texteta\or\texttheta\or\textiota\or\textkappa\or\textlambda
    \or\textmu\or\textnu\or\textxi\or\textpi\or\textsigma
    \or\texttau\or\textupsilon\or\textphi\or\textchi\or\textpsi\or\textomega
    \else\@ctrerr\fi
}

\newcounter{extexpno}
\renewcommand{\theextexpno}{[\greek{extexpno}]}

\newcounter{expno}
\renewcommand{\theexpno}{[\greek{expno}]}
\setcounter{expno}{0}
\begin{table}[t!]\small
\iftoggle{tight}{\vspace{-0.3em}}{}
\begin{center}
\begin{tabular}{@{}llllllllcccccc@{}}\iftoggle{tech_report}{\toprule}{}
Method & SPool & \multicolumn{3}{l}{Image Aug.} & Dim & mAP & \includegraphics[width=0.7cm,height=0.6cm,keepaspectratio]{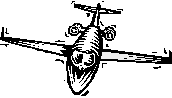} & \includegraphics[width=0.75cm,height=0.6cm,keepaspectratio]{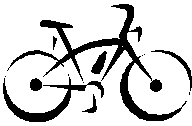} & \includegraphics[width=0.75cm,height=0.6cm,keepaspectratio]{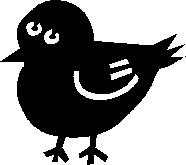} & \includegraphics[width=0.75cm,height=0.6cm,keepaspectratio]{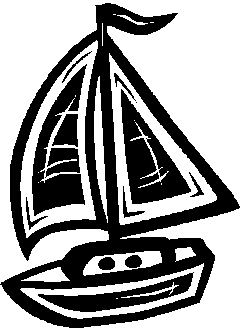} & \includegraphics[width=0.75cm,height=0.6cm,keepaspectratio]{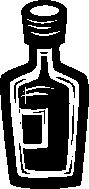} & \includegraphics[width=0.75cm,height=0.6cm,keepaspectratio]{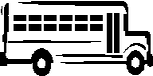}\\
{\small (I) FK BL} & spm & -- & &  & 327K & \textbf{61.69} & 79.0 & 67.4 & 51.9 & 70.9 & 30.8 & 72.2\\
%\label{exp:decaf}{\small (-) DECAF} & -- & -- & &  & 327K & \textbf{72.71} & 87.5 & 78.8 & 83.5 & 77.6 & 42.2 & 72.7\\
\label{exp:decaf_aug_tt}\rowcolor{tableShade}{\small (II) DECAF} & -- & (C) & t & t & 327K & \textbf{73.41} & 87.4 & 79.3 & 84.1 & 78.4 & 42.3 & 73.7\\ \midrule
\refstepcounter{expno}\label{exp:fk}{\small (\greek{expno}) FK} & spm & -- & &  & 327K & \textbf{63.66} & 83.4 & 68.8 & 59.6 & 74.1 & 35.7 & 71.2\\
\refstepcounter{expno}\label{exp:fk_in}\rowcolor{tableShade}{\small (\greek{expno}) FK IN} & spm & -- & &  & 327K & \textbf{64.18} & 82.1 & 69.7 & 59.7 & 75.2 & 35.7 & 71.3\\ \midrule
%\refstepcounter{expno}\label{exp:fk_faboth}\rowcolor{tableShade}{\small (\greek{expno}) FK} & both & -- & &  & 336K & \textbf{64.08} & 84.1 & 69.7 & 59.8 & 74.6 & 35.1 & 71.9\\
\refstepcounter{expno}\label{exp:fk_fa}{\small (\greek{expno}) FK} & (x,y) & -- & &  & 42K & \textbf{63.51} & 83.2 & 69.4 & 60.6 & 73.9 & 36.3 & 68.6\\
\refstepcounter{expno}\label{exp:fk_in_fa}\rowcolor{tableShade}{\small (\greek{expno}) FK IN} & (x,y) & -- & &  & 42K & \textbf{64.36} & 83.1 & 70.4 & 62.4 & 75.2 & 37.1 & 69.1\\
\refstepcounter{expno}\label{exp:fk_in_fa_augfl_fn}{\small (\greek{expno}) FK IN} & (x,y) & (F) & f & - & 42K & \textbf{64.35} & 83.1 & 70.5 & 62.3 & 75.4 & 37.1 & 69.1\\
\refstepcounter{expno}\label{exp:fk_in_fa_aug_fs}\rowcolor{tableShade}{\small (\greek{expno}) FK IN} & (x,y) & (C) & f & s & 42K & \textbf{67.17} & 85.5 & 71.6 & 64.6 & 77.2 & 39.0 & 70.8\\
\refstepcounter{expno}\label{exp:fk_in_fa_aug_ss}{\small (\greek{expno}) FK IN} & (x,y) & (C) & s & s & 42K & \textbf{66.68} & 84.9 & 70.1 & 64.7 & 76.3 & 39.2 & 69.8\\
%\refstepcounter{expno}\label{exp:fk_in_fapsp}{\small (\greek{expno}) FK IN} & (xys) & -- & &  & 42K & \textbf{63.82} & 83.3 & 70.9 & 63.8 & 75.5 & 36.4 & 68.1\\
%\refstepcounter{expno}\label{exp:fk_in_col_fanone}{\small (\greek{expno}) FK IN COL} & -- & -- & &  & 41K & \textbf{50.79} & 69.6 & 52.1 & 46.8 & 62.9 & 24.7 & 48.6\\
%\refstepcounter{expno}\label{exp:fk_in_colplus_fa}{\small (\greek{expno}) FK IN COL+} & (x,y) & -- & &  & 83K & \textbf{65.58} & 83.1 & 70.4 & 66.8 & 74.9 & 35.0 & 69.3\\
%\refstepcounter{expno}\label{exp:fk_in_colplus_fa_aug_fs}{\small (\greek{expno}) FK IN COL+} & (x,y) & (C) & f & s & 83K & \textbf{67.20} & 84.7 & 71.0 & 65.7 & 76.6 & 35.0 & 70.0\\ \midrule
\refstepcounter{expno}\label{exp:fk_in_512_fa}\rowcolor{tableShade}{\small (\greek{expno}) FK IN 512} & (x,y) & -- & &  & 84K & \textbf{65.36} & 84.1 & 70.4 & 65.0 & 76.7 & 37.2 & 71.3\\
\refstepcounter{expno}\label{exp:fk_in_512_fa_aug_fs}{\small (\greek{expno}) FK IN 512} & (x,y) & (C) & f & s & 84K & \textbf{68.02} & 85.9 & 71.8 & 67.1 & 77.1 & 38.8 & 72.3\\
\refstepcounter{expno}\label{exp:fk_in_col_512_fanone}\rowcolor{tableShade}{\small (\greek{expno}) FK IN COL 512} & -- & -- & &  & 82K & \textbf{52.18} & 69.5 & 52.1 & 47.5 & 64.0 & 24.6 & 49.8\\
\refstepcounter{expno}\label{exp:fk_in_512_colplus_fa}{\small (\greek{expno}) FK IN 512 COL+} & (x,y) & -- & &  & 166K & \textbf{66.37} & 82.9 & 70.1 & 67.0 & 77.0 & 36.1 & 70.0\\
\refstepcounter{expno}\label{exp:fk_in_512_colplus_fa_aug_fs}\rowcolor{tableShade}{\small (\greek{expno}) FK IN 512 COL+} & (x,y) & (C) & f & s & 166K & \textbf{67.93} & 85.1 & 70.5 & 67.5 & 77.4 & 35.7 & 71.2\\ \midrule
\refstepcounter{expno}\label{exp:cnn_f_aug_fs}{\small (\greek{expno}) CNN F} & -- & (C) & f & s & 4K & \textbf{77.38} & 88.7 & 83.9 & 87.0 & 84.7 & 46.9 & 77.5\\
\refstepcounter{expno}\label{exp:cnn_s_aug_fs}\rowcolor{tableShade}{\small (\greek{expno}) CNN S} & -- & (C) & f & s & 4K & \textbf{79.74} & 90.7 & 85.7 & 88.9 & 86.6 & 50.5 & 80.1\\ \midrule
\refstepcounter{expno}\label{exp:cnn_m}{\small (\greek{expno}) CNN M} & -- & -- & &  & 4K & \textbf{76.97} & 89.5 & 84.3 & 88.8 & 83.2 & 48.4 & 77.0\\
\refstepcounter{expno}\label{exp:cnn_m_aug_fs}\rowcolor{tableShade}{\small (\greek{expno}) CNN M} & -- & (C) & f & s & 4K & \textbf{79.89} & 91.7 & 85.4 & 89.5 & 86.6 & 51.6 & 79.3\\
\refstepcounter{expno}\label{exp:cnn_m_aug_fm}{\small (\greek{expno}) CNN M} & -- & (C) & f & m & 4K & \textbf{79.50} & 90.9 & 84.6 & 89.4 & 85.8 & 50.3 & 78.4\\
\refstepcounter{expno}\label{exp:cnn_m_aug_ss}\rowcolor{tableShade}{\small (\greek{expno}) CNN M} & -- & (C) & s & s & 4K & \textbf{79.44} & 91.4 & 85.2 & 89.1 & 86.1 & 52.1 & 78.0\\
\refstepcounter{expno}\label{exp:cnn_m_aug_tt}{\small (\greek{expno}) CNN M} & -- & (C) & t & t & 41K & \textbf{78.77} & 90.7 & 85.0 & 89.2 & 85.8 & 51.0 & 77.8\\
\refstepcounter{expno}\label{exp:cnn_m_aug_fn}\rowcolor{tableShade}{\small (\greek{expno}) CNN M} & -- & (C) & f & - & 4K & \textbf{77.78} & 90.5 & 84.3 & 88.8 & 84.5 & 47.9 & 78.0\\
\refstepcounter{expno}\label{exp:cnn_m_augfl_fn}{\small (\greek{expno}) CNN M} & -- & (F) & f & - & 4K & \textbf{76.99} & 90.1 & 84.2 & 89.0 & 83.5 & 48.1 & 77.2\\
\refstepcounter{expno}\label{exp:cnn_m_gs}\rowcolor{tableShade}{\small (\greek{expno}) CNN M GS} & -- & -- & &  & 4K & \textbf{73.59} & 87.4 & 80.8 & 82.4 & 82.1 & 44.5 & 73.5\\
\refstepcounter{expno}\label{exp:cnn_m_gs_aug_fs}{\small (\greek{expno}) CNN M GS} & -- & (C) & f & s & 4K & \textbf{77.00} & 89.4 & 83.8 & 85.1 & 84.4 & 49.4 & 77.6\\ \midrule
\refstepcounter{expno}\label{exp:cnn_m_2048_aug_fs}\rowcolor{tableShade}{\small (\greek{expno}) CNN M 2048} & -- & (C) & f & s & 2K & \textbf{80.10} & 91.3 & 85.8 & 89.9 & 86.7 & 52.4 & 79.7\\
\refstepcounter{expno}\label{exp:cnn_m_1024_aug_fs}{\small (\greek{expno}) CNN M 1024} & -- & (C) & f & s & 1K & \textbf{79.91} & 91.4 & 86.9 & 89.3 & 85.8 & 53.3 & 79.8\\
\refstepcounter{expno}\label{exp:cnn_m_128_aug_fs}\rowcolor{tableShade}{\small (\greek{expno}) CNN M 128} & -- & (C) & f & s & 128 & \textbf{78.60} & 91.3 & 83.9 & 89.2 & 86.9 & 52.1 & 81.0\\ \midrule
\refstepcounter{expno}\label{exp:fkpluscnn_f_fa_aug_fs}{\small (\greek{expno}) FK+CNN F} & (x,y) & (C) & f & s & 88K & \textbf{77.95} & 89.6 & 83.1 & 87.1 & 84.5 & 48.0 & 79.4\\
%\refstepcounter{expno}\label{exp:fkpluscnn_m_fa_aug_fs}{\small (\greek{expno}) FK+CNN M} & (x,y) & (C) & f & s & 88K & \textbf{79.77} & 90.9 & 84.5 & 88.7 & 85.0 & 52.2 & 80.3\\
\refstepcounter{expno}\label{exp:fkpluscnn_m_2048_fa_aug_fs}\rowcolor{tableShade}{\small (\greek{expno}) FK+CNN M 2048} & (x,y) & (C) & f & s & 86K & \textbf{80.14} & 90.9 & 85.9 & 88.8 & 85.5 & 52.3 & 81.4\\ \midrule
\refstepcounter{expno}\label{exp:cnn_s_aug_fs_tune}{\small (\greek{expno}) CNN S TUNE-RNK} & -- & (C) & f & s & 4K & \textbf{82.42} & 95.3 & 90.4 & 92.5 & 89.6 & 54.4 & 81.9\\ \bottomrule
\end{tabular}
\end{center}
\iftoggle{tight}{\vspace{-1em}}{}
\caption{VOC 2007 results \emph{(continued overleaf)}. See Sect.~\ref{sec:analysis} for details.}\label{tab:res_tab}
\iftoggle{tight}{\vspace{-0.4em}}{}
\end{table}

\setcounter{expno}{0}
\begin{table}[ht!]\small
\iftoggle{tight}{\vspace{-0.3em}}{}
\begin{center}
\begin{tabular}{@{}lcccccccccccccc@{}}\iftoggle{tech_report}{\toprule}{}
 & \includegraphics[width=0.75cm,height=0.6cm,keepaspectratio]{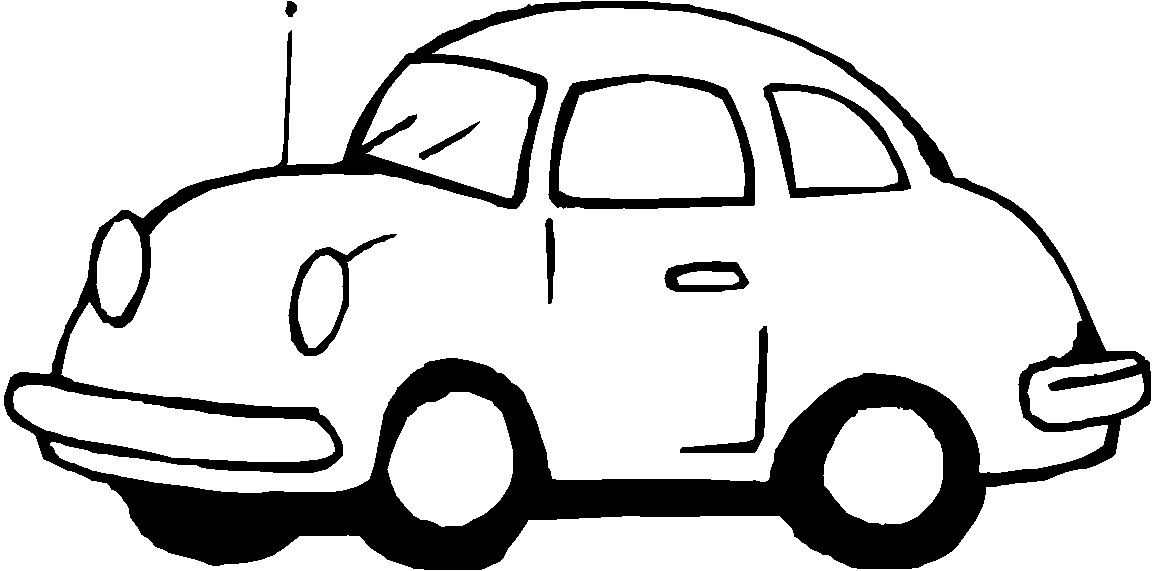} & \includegraphics[width=0.75cm,height=0.6cm,keepaspectratio]{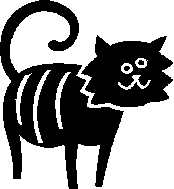} & \includegraphics[width=0.75cm,height=0.6cm,keepaspectratio]{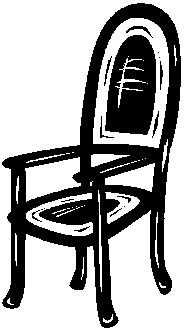} & \includegraphics[width=0.75cm,height=0.6cm,keepaspectratio]{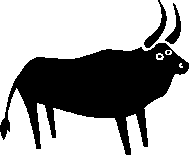} & \includegraphics[width=0.75cm,height=0.6cm,keepaspectratio]{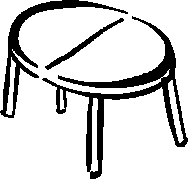} & \includegraphics[width=0.75cm,height=0.6cm,keepaspectratio]{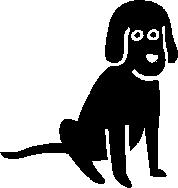} & \includegraphics[width=0.75cm,height=0.6cm,keepaspectratio]{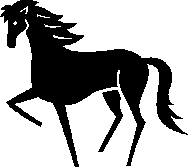} & \includegraphics[width=0.75cm,height=0.6cm,keepaspectratio]{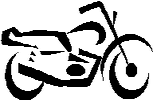} & \includegraphics[width=0.75cm,height=0.6cm,keepaspectratio]{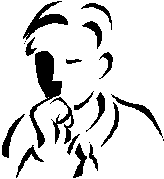} & \includegraphics[width=0.75cm,height=0.6cm,keepaspectratio]{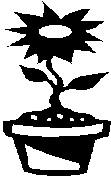} & \includegraphics[width=0.75cm,height=0.6cm,keepaspectratio]{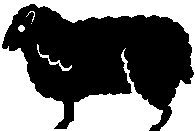} & \includegraphics[width=0.75cm,height=0.6cm,keepaspectratio]{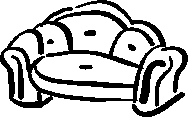} & \includegraphics[width=0.75cm,height=0.6cm,keepaspectratio]{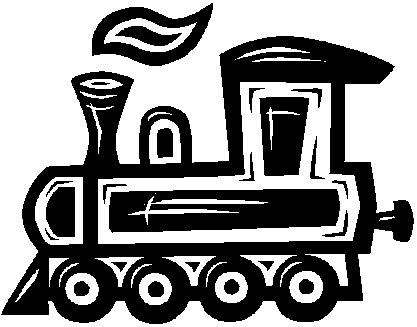} & \includegraphics[width=0.75cm,height=0.6cm,keepaspectratio]{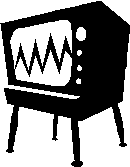}\\
{\small (I)} & 79.9 & 61.4 & 56.0 & 49.6 & 58.4 & 44.8 & 78.8 & 70.8 & 85.0 & 31.7 & 51.0 & 56.4 & 80.2 & 57.5\\
%{\small (-)} & 83.5 & 83.1 & 53.9 & 61.4 & 68.6 & 78.5 & 84.1 & 75.6 & 90.4 & 49.5 & 74.2 & 56.3 & 85.9 & 66.8\\
\rowcolor{tableShade}{\small (II)} & 83.7 & 83.7 & 54.3 & 61.9 & 70.2 & 79.5 & 85.3 & 77.2 & 90.9 & 51.1 & 73.8 & 57.0 & 86.4 & 68.0\\ \midrule
\refstepcounter{expno}{\small (\greek{expno})} & 80.7 & 64.4 & 53.8 & 53.8 & 60.2 & 47.8 & 79.9 & 68.9 & 86.1 & 37.3 & 51.1 & 55.8 & 83.7 & 56.9\\
\refstepcounter{expno}\rowcolor{tableShade}{\small (\greek{expno})} & 80.6 & 64.8 & 53.9 & 54.9 & 60.7 & 50.5 & 80.4 & 69.5 & 86.2 & 38.3 & 54.4 & 56.3 & 82.7 & 56.7\\ \midrule
%\refstepcounter{expno}\rowcolor{tableShade}{\small (\greek{expno})} & 80.6 & 64.8 & 54.4 & 53.5 & 62.9 & 48.9 & 80.3 & 68.5 & 86.0 & 36.7 & 52.4 & 55.8 & 83.4 & 58.1\\
\refstepcounter{expno}{\small (\greek{expno})} & 81.1 & 64.2 & 51.1 & 53.4 & 61.9 & 50.0 & 80.0 & 67.5 & 85.3 & 35.7 & 51.9 & 53.8 & 83.5 & 58.9\\
\refstepcounter{expno}\rowcolor{tableShade}{\small (\greek{expno})} & 80.5 & 66.9 & 50.9 & 53.9 & 62.1 & 51.5 & 80.5 & 68.5 & 85.9 & 37.2 & 55.2 & 54.3 & 83.3 & 59.2\\
\refstepcounter{expno}{\small (\greek{expno})} & 80.5 & 66.8 & 51.0 & 54.1 & 62.2 & 51.5 & 80.4 & 68.2 & 86.0 & 37.3 & 55.1 & 54.2 & 83.3 & 59.2\\
\refstepcounter{expno}\rowcolor{tableShade}{\small (\greek{expno})} & 82.4 & 71.6 & 52.8 & 62.4 & 63.4 & 57.1 & 81.6 & 70.9 & 86.9 & 41.2 & 61.2 & 56.9 & 85.2 & 61.5\\
\refstepcounter{expno}{\small (\greek{expno})} & 81.9 & 71.0 & 52.8 & 61.6 & 62.2 & 56.8 & 81.8 & 70.0 & 86.5 & 41.5 & 61.0 & 56.5 & 84.3 & 60.9\\
%\refstepcounter{expno}{\small (\greek{expno})} & 80.0 & 67.1 & 51.8 & 50.9 & 59.3 & 51.5 & 79.7 & 68.5 & 85.9 & 35.1 & 54.5 & 53.3 & 83.8 & 57.1\\
%\refstepcounter{expno}{\small (\greek{expno})} & 64.0 & 44.1 & 41.9 & 34.3 & 35.5 & 42.3 & 74.6 & 58.0 & 82.9 & 39.0 & 45.1 & 35.9 & 65.0 & 48.7\\
%\refstepcounter{expno}{\small (\greek{expno})} & 79.6 & 66.2 & 51.6 & 53.7 & 59.3 & 55.1 & 80.9 & 69.9 & 87.7 & 45.4 & 59.7 & 55.8 & 84.7 & 62.3\\
%\refstepcounter{expno}{\small (\greek{expno})} & 80.9 & 70.6 & 51.4 & 58.4 & 62.6 & 58.8 & 82.2 & 70.8 & 88.3 & 49.4 & 61.2 & 56.6 & 86.1 & 63.8\\ \midrule
\refstepcounter{expno}\rowcolor{tableShade}{\small (\greek{expno})} & 81.1 & 67.9 & 52.6 & 55.4 & 61.4 & 51.2 & 80.5 & 69.1 & 86.4 & 41.2 & 56.0 & 56.2 & 83.7 & 59.9\\
\refstepcounter{expno}{\small (\greek{expno})} & 82.5 & 73.2 & 54.7 & 62.7 & 64.5 & 56.6 & 82.2 & 71.3 & 87.5 & 43.0 & 62.0 & 59.3 & 85.7 & 62.4\\
\refstepcounter{expno}\rowcolor{tableShade}{\small (\greek{expno})} & 66.1 & 46.6 & 42.5 & 35.8 & 41.1 & 45.5 & 75.4 & 58.3 & 83.9 & 39.8 & 47.3 & 35.6 & 69.2 & 49.0\\
\refstepcounter{expno}{\small (\greek{expno})} & 80.0 & 65.9 & 52.8 & 56.1 & 61.0 & 56.9 & 81.4 & 69.6 & 88.4 & 49.0 & 59.2 & 56.4 & 84.7 & 62.8\\
\refstepcounter{expno}\rowcolor{tableShade}{\small (\greek{expno})} & 81.6 & 70.8 & 52.9 & 59.6 & 63.1 & 59.9 & 82.1 & 70.5 & 88.9 & 50.6 & 63.7 & 57.5 & 86.1 & 64.1\\ \midrule
\refstepcounter{expno}{\small (\greek{expno})} & 86.3 & 85.4 & 58.6 & 71.0 & 72.6 & 82.0 & 87.9 & 80.7 & 91.8 & 58.5 & 77.4 & 66.3 & 89.1 & 71.3\\
\refstepcounter{expno}\rowcolor{tableShade}{\small (\greek{expno})} & 87.8 & 88.3 & 61.3 & 74.8 & 74.7 & 87.2 & 89.0 & 83.7 & 92.3 & 58.8 & 80.5 & 69.4 & 90.5 & 74.0\\ \midrule
\refstepcounter{expno}{\small (\greek{expno})} & 85.1 & 87.4 & 58.1 & 70.4 & 73.1 & 83.5 & 85.5 & 80.9 & 90.8 & 54.1 & 78.9 & 61.1 & 89.0 & 70.4\\
\refstepcounter{expno}\rowcolor{tableShade}{\small (\greek{expno})} & 87.7 & 88.6 & 60.3 & 80.1 & 74.4 & 85.9 & 88.2 & 84.6 & 92.1 & 60.3 & 80.5 & 66.2 & 91.3 & 73.5\\
\refstepcounter{expno}{\small (\greek{expno})} & 87.6 & 88.6 & 60.7 & 78.2 & 73.6 & 86.0 & 87.4 & 83.8 & 92.3 & 59.3 & 81.0 & 66.8 & 91.3 & 74.0\\
\refstepcounter{expno}\rowcolor{tableShade}{\small (\greek{expno})} & 87.5 & 88.1 & 60.4 & 76.9 & 74.8 & 85.8 & 88.1 & 84.3 & 92.2 & 59.5 & 79.3 & 65.8 & 90.8 & 73.5\\
\refstepcounter{expno}{\small (\greek{expno})} & 87.3 & 87.6 & 60.1 & 72.3 & 75.3 & 85.2 & 86.9 & 82.6 & 91.9 & 58.5 & 77.9 & 66.5 & 90.5 & 73.4\\
\refstepcounter{expno}\rowcolor{tableShade}{\small (\greek{expno})} & 85.7 & 87.9 & 58.3 & 74.2 & 73.9 & 84.7 & 86.6 & 82.0 & 91.0 & 55.8 & 79.2 & 62.1 & 89.3 & 71.0\\
\refstepcounter{expno}{\small (\greek{expno})} & 85.3 & 87.3 & 58.1 & 70.0 & 73.4 & 83.5 & 86.0 & 80.8 & 90.9 & 53.9 & 78.1 & 61.2 & 88.8 & 70.6\\
\refstepcounter{expno}\rowcolor{tableShade}{\small (\greek{expno})} & 85.0 & 84.9 & 57.8 & 65.9 & 69.8 & 79.5 & 82.9 & 77.4 & 89.2 & 42.8 & 71.7 & 60.2 & 86.3 & 67.8\\
\refstepcounter{expno}{\small (\greek{expno})} & 87.2 & 86.5 & 59.5 & 72.4 & 74.1 & 81.7 & 86.0 & 82.3 & 90.8 & 48.9 & 73.7 & 66.8 & 89.6 & 71.0\\ \midrule
\refstepcounter{expno}\rowcolor{tableShade}{\small (\greek{expno})} & 87.6 & 88.4 & 60.2 & 76.9 & 75.4 & 85.5 & 88.0 & 83.4 & 92.1 & 61.1 & 83.1 & 68.5 & 91.9 & 74.2\\
\refstepcounter{expno}{\small (\greek{expno})} & 87.8 & 88.6 & 59.0 & 77.2 & 73.1 & 85.9 & 88.3 & 83.5 & 91.8 & 59.9 & 81.4 & 68.3 & 93.0 & 74.1\\
\refstepcounter{expno}\rowcolor{tableShade}{\small (\greek{expno})} & 86.6 & 87.5 & 59.1 & 70.0 & 72.9 & 84.6 & 86.7 & 83.6 & 89.4 & 57.0 & 81.5 & 64.8 & 90.4 & 73.4\\ \midrule
\refstepcounter{expno}{\small (\greek{expno})} & 86.8 & 85.6 & 59.9 & 72.0 & 73.4 & 81.4 & 88.6 & 80.5 & 92.1 & 60.6 & 77.3 & 66.4 & 89.3 & 73.3\\
%\refstepcounter{expno}{\small (\greek{expno})} & 87.7 & 88.6 & 60.5 & 77.4 & 76.0 & 84.8 & 88.9 & 83.6 & 92.4 & 60.8 & 79.9 & 67.2 & 91.1 & 74.8\\
\refstepcounter{expno}\rowcolor{tableShade}{\small (\greek{expno})} & 87.7 & 88.4 & 61.2 & 76.9 & 76.6 & 84.9 & 89.1 & 82.9 & 92.4 & 61.9 & 80.9 & 68.7 & 91.5 & 75.1\\ \midrule
\refstepcounter{expno}{\small (\greek{expno})} & 91.5 & 91.9 & 64.1 & 76.3 & 74.9 & 89.7 & 92.2 & 86.9 & 95.2 & 60.7 & 82.9 & 68.0 & 95.5 & 74.4\\ \bottomrule
\end{tabular}
\end{center}
\iftoggle{tight}{\vspace{-1em}}{}
\addtocounter{table}{-1}
\caption{VOC 2007 results \emph{(continued from previous page)}}
\iftoggle{tight}{\vspace{-0.7em}}{}
\end{table}

%{\small (I) FK BL} & spm & -- & &  & 327K & \textbf{61.69} & 79.0 & 67.4 & 51.9 & 70.9 & 30.8 & 72.2\\
%
%{\small (I)} & 79.9 & 61.4 & 56.0 & 49.6 & 58.4 & 44.8 & 78.8 & 70.8 & 85.0 & 31.7 & 51.0 & 56.4 & 80.2 & 57.5\\

%\rk{Removed results from table: work into text instead (spatial feature augmentation with scale gives 63.82 vs 64.36, decaf result is 72.71/73.41 w/wo aug}

This section describes the experimental results, comparing different features and data augmentation schemes.
%In all cases, features $\enc(\image)$ are
%computed for each image $\image$ as either the output of the IFV (the
%shallow encoding method described in Sect.~\ref{sec:ifv}), or the penultimate
%layer of a CNN network, as described in Sect~\ref{sec:cnn}.
The results are given in~\tblref{tab:res_tab} for VOC-2007 and analysed next, starting from generally applicable methods such as augmentation and then discussing the specifics of each scenario.
We then move onto other datasets and the state of the art in~\sref{sec:state_of_art}.

%a detailed
%%analysis follows, starting by evaluating those aspects of %the pipeline which can
%be applied across both shallow and deep methods, assessing %their applicability
%to both, and then considering points relating to each of %the individual
%experimental scenarios in turn.

\iftoggle{tight}{\vspace{-0.2em}}{}
\subsectionpar{Data augmentation} We experiment with no data
augmentation (denoted \emph{Image Aug=--} in Tab.~\ref{tab:res_tab}),
flip augmentation (\emph{Image Aug=F}), and C+F augmentation
(\emph{Image Aug=C}). Augmented images are used as stand-alone
samples (\emph{f}), or by fusing the corresponding descriptors using
sum (\emph{s}) or max (\emph{m}) pooling or
stacking (\emph{t}). So for example $\emph{Image Aug=(C) f s}$ in row
\ref{exp:fk_in_fa_aug_fs} of Tab.~\ref{tab:res_tab} means that C+F augmentation is
used to generate additional samples in training (\emph{f}), and is combined with sum-pooling
in testing ($\emph{s}$).

Augmentation consistently improves performance by $\sim3\%$ for both IFV
(\eg\ref{exp:fk_in_fa} \vs\ref{exp:fk_in_fa_aug_fs}) and CNN
(\eg\ref{exp:cnn_m} \vs\ref{exp:cnn_m_aug_fs}).
Using additional samples for training and sum-pooling for testing works
best (\ref{exp:cnn_m_aug_fs}) followed by sum-pooling \ref{exp:cnn_m_aug_ss}, max pooling \ref{exp:cnn_m_aug_fm}, and stacking \ref{exp:cnn_m_aug_tt}.
In terms of the choice of transformations, flipping improves only marginally (\ref{exp:cnn_m} \vs\ref{exp:cnn_m_augfl_fn}), but using the more expensive C+F sampling improves, as seen, by about $2\sim3\%$ (\ref{exp:cnn_m} \vs\ref{exp:cnn_m_aug_fs}). We experimented with sampling more transformations,
taking a higher density of crops from the centre of the image, but observed no benefit.

% We experimented with sampling more transformations but observed no benefit~(\ref{exp:cnn_m_grid5_fs}).

% both for the Fisher encoding (\ref{exp:fk_in_fa}
%vs. \ref{exp:fk_in_fa_aug_fs}) and CNN-based methods (\ref{exp:cnn_m}
%vs. \ref{exp:cnn_m_aug_fs}). There appears to be a limit to the gains
%to be made by increasing data sampling though, as when we sample with
%G=3$ (giving a total of 26 images rather than the standard 10) no
%further performance increase is observed \ref{exp:cnn_m_grid5_fs}.

\iftoggle{tight}{\vspace{-0.2em}}{}
\subsectionpar{Colour} Colour information can be added and subtracted in
CNN and IFV. In IFV replacing SIFT with the colour descriptors of~\cite{Perronnin10a} (denoted
\emph{COL} in \emph{Method}) yields
significantly worse performance (\ref{exp:fk_in_col_512_fanone} \vs\ref{exp:fk_in_512_fa}). However, when SIFT and colour descriptors are
combined by stacking the corresponding IFVs (\emph{COL+}) there is a small but significant improvement of around $\sim 1\%$ in the non-augmented case (\eg\ref{exp:fk_in_512_fa} \vs\ref{exp:fk_in_512_colplus_fa}) but little impact in
the augmented case (\eg\ref{exp:fk_in_512_fa_aug_fs} \vs\ref{exp:fk_in_512_colplus_fa_aug_fs}). For CNNs, retraining the network after converting all the input images to grayscale
(denoted \emph{GS} in \emph{Methods}) has a more significant impact, resulting
in a performance drop of $\sim 3\%$ (\ref{exp:cnn_m_gs_aug_fs}
\vs\ref{exp:cnn_m_aug_fs}, \ref{exp:cnn_m_gs} \vs\ref{exp:cnn_m}).

%\begin{wrapfigure}[8]{r}[10pt]{6.5cm}
%\iftoggle{tight}{\vspace{-1.5em}}{}
%\begin{tabular}{@{}lccc@{}}\toprule
%Method & Prenorm & Postnorm & mAP\\
%CNN-M  & --      & L2       & 79.78\\
%       & L2      & L2       & 79.73\\
%       & L2      & --       & 74.91\\
%FK-IN  & --      & L2       & 67.17\\
%       & L2      & L2       & 67.06\\
%       & L2      & --       & 67.08\\ \bottomrule
%\end{tabular}
%\end{wrapfigure}

\iftoggle{tight}{\vspace{-0.2em}}{}
\subsectionpar{Scenario 1: Shallow representation (IFV)} The baseline IFV
encoding using a spatial pyramid \ref{exp:fk} performs slightly better
than the results~[I] taken from
Chatfield~\etal~\cite{Chatfield11}, primarily due to a larger number
of spatial scales being used during SIFT feature extraction, and the
resultant SIFT features being
square-rooted. \emph{Intra-normalisation}, denoted as \emph{IN} in the
\emph{Method} column of the table,
improves the performance by $\sim 1\%$ (\eg \ref{exp:fk_fa} \vs\ref{exp:fk_in_fa}). More interestingly,
switching from spatial pooling (denoted \emph{spm} in the \emph{SPool}
column) to feature spatial augmentation
(\emph{SPool=(x,y)}) has either little effect on
the performance or results in a marginal increase (\ref{exp:fk}
\vs\ref{exp:fk_fa}, \ref{exp:fk_in} \vs\ref{exp:fk_in_fa}), whilst
resulting in a representation which is over 10$\times$
smaller. We also experimented with augmenting with scale in addition to position
as in~\cite{Sanchez12} but observed no improvement.
Finally, we investigate pushing the parameters of the
representation setting $K=512$ (rows~\ref{exp:fk_in_512_fa}-\ref{exp:fk_in_512_colplus_fa_aug_fs}). Increasing the number of
GMM centres in the model from $K=256$ to $512$ results in a further
performance increase (e.g.\ \ref{exp:fk_in_512_fa} \vs\ref{exp:fk_in_fa}),
but at the expense of higher-dimensional codes (125K dimensional).

\iftoggle{tight}{\vspace{-0.2em}}{}
\subsectionpar{Scenario 2: Deep representation (CNN) with pre-training}
CNN-based methods consistently outperform the shallow encodings, even after the improvements discussed above, by a large $\sim 10\%$ mAP margin (\ref{exp:fk_in_512_fa_aug_fs} \vs\ref{exp:cnn_m_aug_fs}). Our small architecture CNN-F, which is similar to DeCAF~\cite{Donahue13}, performs significantly better than the latter ([II] \vs\ref{exp:cnn_m_aug_tt}), validating our implementation.
Both medium CNN-M~\ref{exp:cnn_f_aug_fs} and slow CNN-S~\ref{exp:cnn_m_aug_fs} outperform  the fast CNN-F~\ref{exp:cnn_f_aug_fs} by a significant $2 \sim 3 \%$ margin. Since the accuracy of CNN-S and CNN-M is nearly the same, we focus on the latter as it is simpler and  marginally ($\sim 25\%$) faster. Remarkably, these good networks work very well even with no augmentation~\ref{exp:cnn_m}. Another advantage of CNNs compared to IFV is the small dimensionality of the output features, although IFV can be compressed to an extent. We explored retraining the CNNs such
that the final layer was of a lower dimensionality, and reducing from 4096 to
2048 actually resulted in a marginal performance boost (\ref{exp:cnn_m_2048_aug_fs} \vs\ref{exp:cnn_m_aug_fs}). What is surprising is that we can reduce the output
dimensionality further to 1024D~\ref{exp:cnn_m_1024_aug_fs} and even 128D~\ref{exp:cnn_m_128_aug_fs} with only a drop of $\sim2 \%$ for codes
that are $32 \times$ smaller ($\sim650 \times$ smaller than our best performing IFV~\ref{exp:fk_in_512_fa_aug_fs}). Note, $\ell^2$-normalising the features accounted for up to $\sim 5\%$ of their performance over VOC 2007; it %normalisation
should be applied before input to the SVM and after pooling the augmented descriptors (where applicable).

\iftoggle{tight}{\vspace{-0.2em}}{}
\subsectionpar{Scenario 3: Deep representation (CNN) with pre-training and fine-tuning}
We fine-tuned our CNN-S architecture on VOC-2007 using the ranking hinge loss, and achieved a significant improvement: $2.7\%$ (\ref{exp:cnn_s_aug_fs_tune} \vs\ref{exp:cnn_s_aug_fs}).
This demonstrates that in spite of the small amount of VOC training data (5,011 images), fine-tuning is able to adjust the learnt
deep representation to better suit the dataset in question.

\iftoggle{tight}{\vspace{-0.2em}}{}
\subsectionpar{Combinations} For the CNN-M 2048 representation~\ref{exp:cnn_m_2048_aug_fs}, stacking deep and shallow representations to form
a higher-dimensional descriptor makes little difference (\ref{exp:cnn_m_2048_aug_fs} \vs\ref{exp:fkpluscnn_m_2048_fa_aug_fs}). For
the weaker CNN-F it results in a small boost of $\sim 0.8\%$ (\ref{exp:cnn_f_aug_fs} \vs\ref{exp:fkpluscnn_f_fa_aug_fs}).

\begin{table}[t!]
\small
\begin{center}
\setcounter{extexpno}{0}
%\setlength{\tabcolsep}{3pt}
%\renewcommand{\arraystretch}{0.83}% reduce row height
% \begin{wrapfigure}[5]{r}[10pt]{6cm}
\iftoggle{tight}{\vspace{-0.3em}}{}
\begin{tabular}{@{}lccccc@{}}\iftoggle{tech_report}{\toprule}{}
 & \textbf{ILSVRC-2012} & \textbf{VOC-2007} & \textbf{VOC-2012} & \textbf{Caltech-101} & \textbf{Caltech-256} \\
 & (top-5 error) & (mAP) & (mAP) & (accuracy) & (accuracy) \\
\refstepcounter{extexpno}(\greek{extexpno}) FK IN 512 & - & 68.0 & -- & -- & -- \\ \midrule
%\refstepcounter{extexpno}(\greek{extexpno}) FK IN 512 COL+ & - & 67.9 & -- & -- \\ \midrule
\refstepcounter{extexpno}\rowcolor{tableShade}(\greek{extexpno}) CNN F & 16.7 & 77.4 & 79.9 & -- & -- \\
\refstepcounter{extexpno}(\greek{extexpno}) CNN M & 13.7 & 79.9 & 82.5 & 87.15 $\pm$ 0.80 & 77.03 $\pm$ 0.46\\
\refstepcounter{extexpno}\rowcolor{tableShade}(\greek{extexpno}) CNN M 2048 & 13.5 & 80.1 & 82.4 & 86.64 $\pm$ 0.53 & 76.88 $\pm$ 0.35\\
\refstepcounter{extexpno}(\greek{extexpno}) CNN S & \textbf{13.1} & 79.7 & 82.9 & 87.76 $\pm$ 0.66 & \textbf{77.61 $\pm$ 0.12} \\ 
\refstepcounter{extexpno}\label{exp:final_cnn_s_tune_cls}\rowcolor{tableShade}(\greek{extexpno}) CNN S TUNE-CLS & \textbf{13.1} & - & 83.0 & \textbf{88.35 $\pm$ 0.56} & 77.33 $\pm$ 0.56 \\
\refstepcounter{extexpno}\label{exp:final_cnn_s_tune_rnk}(\greek{extexpno}) CNN S TUNE-RNK & \textbf{13.1} & \textbf{82.4} & \textbf{83.2}  & -- & --\\ \midrule \midrule
\refstepcounter{extexpno}\label{exp:final_zf}\rowcolor{tableShade}(\greek{extexpno}) Zeiler \& Fergus~\cite{Zeiler13} & 16.1 & - & 79.0 & 86.5 $\pm$ 0.5 & 74.2 $\pm$ 0.3 \\
\refstepcounter{extexpno}(\greek{extexpno}) Razavian~\etal~\cite{Razavian14,Sermanet14} & 14.7 & 77.2 & -- & -- & --\\ 
\refstepcounter{extexpno}\label{exp:final_oquab}\rowcolor{tableShade}(\greek{extexpno}) Oquab~\etal\cite{Oquab14} & 18 & 77.7 & 78.7 (82.8\textsuperscript{*}) & -- & --\\ 
\refstepcounter{extexpno}\label{exp:final_oquab_nips}(\greek{extexpno}) Oquab~\etal\cite{Oquab14a} & - & - & \textbf{86.3\textsuperscript{*}} & -- & --\\ 
\refstepcounter{extexpno}\label{exp:final_wei}\rowcolor{tableShade}(\greek{extexpno}) Wei~\etal\cite{Wei14} & - & 81.5 (\textbf{85.2\textsuperscript{*}}) & 81.7 (\textbf{90.3\textsuperscript{*}}) & -- & --\\
\refstepcounter{extexpno}\label{exp:final_he}(\greek{extexpno}) He~\etal\cite{He14} & 13.6 & 80.1 & - & \textbf{91.4 $\pm$ 0.7} & --\\ \bottomrule
\end{tabular}
\end{center}
\iftoggle{tight}{\vspace{-1em}}{}
\caption{\textbf{Comparison with the state of the art} on ILSVRC2012, VOC2007, VOC2012, Caltech-101, and Caltech-256.
Results marked with * were achieved using models pre-trained on the \emph{extended} ILSVRC datasets (1512 classes in~\cite{Oquab14,Oquab14a}, 2000 classes in~\cite{Wei14}).
All other results were achieved using CNNs pre-trained on ILSVRC-2012 (1000 classes).
}
\label{tab:voc12_ilsvrc}
\iftoggle{tight}{\vspace{-0.5em}}{}
% \end{wrapfigure}
\end{table}

\iftoggle{tight}{\vspace{-0.2em}}{}
\subsectionpar{Comparison with the state of the art} 
\label{sec:state_of_art}
In~\tblref{tab:voc12_ilsvrc} we report our results on ILSVRC-2012, VOC-2007, VOC-2012, Caltech-101, and Caltech-256 datasets, and compare them to the state of the art. 
First, we note that the ILSVRC error rates of our CNN-F, CNN-M, and CNN-S networks are better than those reported by~\cite{Krizhevsky12},~\cite{Zeiler13},
and~\cite{Sermanet14} for the related configurations. This validates our implementation, and the difference is likely
to be due to the sampling of image crops from the uncropped image plane (instead of the centre).
When using our CNN features on other datasets, the relative performance generally follows the same pattern as on ILSVRC, where the nets are trained -- 
the CNN-F architecture exhibits the worst performance, with CNN-M and CNN-S performing considerably better.

Further fine-tuning of CNN-S on the VOC datasets turns out to be beneficial; on VOC-2012, using the ranking loss is marginally better
than the classification loss (\ref{exp:final_cnn_s_tune_rnk} \vs \ref{exp:final_cnn_s_tune_cls}), which can be explained by the ranking-based VOC 
evaluation criterion. Fine-tuning on Caltech-101 also yields a small improvement, but
no gain is observed over Caltech-256.
%Interestingly, unlike the VOC datasets, fine-tuning on Caltech-101 did not improve the performance.
%This can be explained by the limited amount of training data ($\sim$30 samples / class). While this is sufficient to train 
%an SVM, it might not be enough to adjust the weights of a deep architecture.

Our CNN-S net is competitive with recent CNN-based approaches~\cite{Zeiler13,Razavian14,Oquab14,Oquab14a,Wei14,He14} and on a number of datasets (VOC-2007, VOC-2012, Caltech-101, Caltech-256)
and sets the state of the art on VOC-2007 and VOC-2012 across methods pre-trained solely on ILSVRC-2012 dataset.
While the CNN-based methods of~\cite{Oquab14a,Wei14} achieve better performance on VOC (86.3\% and 90.3\% respectively), they were trained using extended ILSVRC datasets, enriched with additional 
categories semantically close to the ones in VOC.
Additionally,~\cite{Wei14} used a significantly more complex classification pipeline, driven by bounding box proposals~\cite{Cheng14}, pre-trained on ILSVRC-2013 detection dataset.
Their best reported result on VOC-2012 (90.3\%) was achieved by the late fusion with a complex hand-crafted method of~\cite{Yan12a}; without fusion, they get 84.2\%.
% Interestingly, on the VOC datasets we achieve superior results to~\cite{Oquab14} (\ref{exp:final_oquab}) with a conceptually simpler method, but with a more powerful deep architecture.
% In their case, they exploited ground-truth object bounding boxes, additional VOC-like ImageNet categories, and a large number of multi-scale crops, but used 
% a less accurate CNN. Without extra categories (\ie by using only ILSVRC data as we do), they report 78.7\% on VOC-2012.
% The CNN architecture of~\cite{Zeiler13} (\ref{exp:final_zf}) is similar to ours, but their use of softmax regression loss is not 
% well-suited to multi-label VOC data.
On Caltech-101,~\cite{He14} achieves the state of the art using spatial pyramid pooling of conv5 layer features, while we used full7 layer features consistently
across all datasets (for full7 features, they report $87.08\%$).

In addition to achieving performance comparable to the state of the art with a very simple approach
(but powerful CNN-based features), with the modifications outlined in the paper (primarily the
use of data augmentation similar to the CNN-based methods) we are able to improve the performance of shallow IFV to 68.02\% (\tblref{tab:res_tab}, \ref{exp:fk_in_512_fa_aug_fs}).

\begin{figure}[t]
    \centering
    \includegraphics[width=0.8\textwidth]{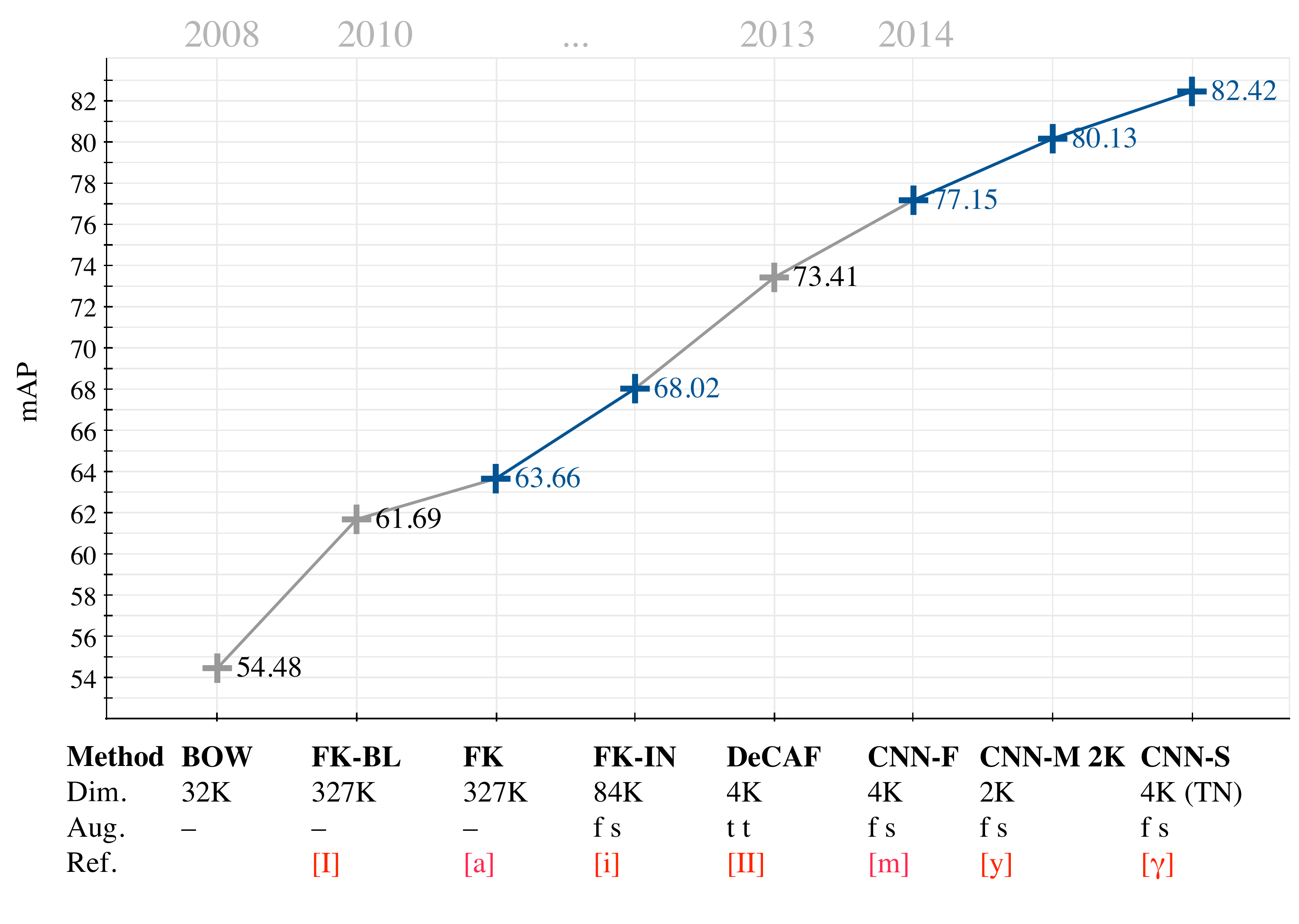}
    \caption{\textbf{Evolution of Performance on PASCAL VOC-2007 over the recent years.} 
    Please refer to~\tblref{tab:res_tab} for details and references.}
    \label{fig:vocevo}
\end{figure}

\subsectionpar{Performance Evolution on VOC-2007}
A comparative plot of the evolution in the performance of the methods evaluated in
this paper, along with a selection from our earlier review of shallow methods~\cite{Chatfield11} is presented in~\figref{fig:vocevo}.
Classification accuracy over PASCAL VOC was 54.48\% mAP
for the BoVW model in 2008, 61.7\% for the IFV in 2010~\cite{Chatfield11}, and
73.41\% for DeCAF~\cite{Donahue13} and similar~\cite{Oquab14,Razavian14} CNN-based
methods introduced in late 2013. Our best performing CNN-based method (CNN-S with
fine-tuning) achieves 82.42\%, comparable to the most recent state-of-the-art.

\iftoggle{tight}{\vspace{-0.2em}}{}
\subsectionpar{Timings and dimensionality} One of our best-performing CNN representations CNN-M-2048~\ref{exp:cnn_m_2048_aug_fs} is $\sim42\times$ more compact than the best performing
IFV~\ref{exp:fk_in_512_fa_aug_fs} (84K vs.~2K) and CNN-M features are also $\sim50\times$ faster to compute ($\sim120s$ vs.~$\sim2.4s$ per image with augmentation enabled, over a single CPU core).
Non-augmented CNN-M features~\ref{exp:cnn_m} take around $0.3s$ per image, compared to $\sim0.4s$ for CNN-S features and $\sim0.13s$ for CNN-F features.

%------------------------------------------------------------------------- 
\iftoggle{tight}{\vspace{-0.5em}}{}
\section{Conclusion}
\label{sec:conc}
\iftoggle{tight}{\vspace{-0.3em}}{}

In this paper we presented a rigorous empirical evaluation of CNN-based methods for image classification, along with a comparison with more traditional
shallow feature encoding methods. We have demonstrated that the performance of shallow representations can be significantly improved by adopting data augmentation, 
typically used in deep learning. In spite of this improvement, deep architectures still outperform the shallow methods by a large margin.
We have shown that the performance of deep representations on the ILSVRC dataset is a good indicator of their performance on other datasets, and that fine-tuning
can further improve on already very strong results achieved using the combination of deep representations and a linear SVM.
Source code and CNN models to reproduce the experiments presented in the paper are available on the project website~\cite{deepevalpage} in the hope that it would
provide common ground for future comparisons, and good baselines for image representation
research.% In future work, we will further explore the space of the CNN configurations.

%-------------------------------------------------------------------------

\iftoggle{tight}{\vspace{-0.2em}}{}
\iftoggle{tech_report}{%
\section*{Acknowledgements}
}{%
\paragraph{Acknowledgements.}%
}
This work was supported by the EPSRC and ERC grant VisRec no. 228180.
We gratefully acknowledge the support
of NVIDIA Corporation with the donation of the GPUs used for this research.
%Thanks to NVIDIA Corporation for the donation of the GPUs used for this research.

\iftoggle{tight}{%
\input{compactify_end.tex}
}{}

\iftoggle{tech_report}{%
\bibliographystyle{IEEEtran}
}{}

\bibliography{bib/longstrings,bib/shortstrings,bib/vgg_local,bib/vgg_other,bib/vgg_devil}
\end{document}

%% file: abstract.tex
The latest generation of Convolutional Neural Networks (CNN) have achieved
impressive results in challenging benchmarks on image recognition and object
detection, significantly raising the interest of the community in these
methods. Nevertheless, it is still unclear how different CNN methods compare
with each other and with previous state-of-the-art shallow representations such
as the Bag-of-Visual-Words and the Improved Fisher Vector. This paper conducts a
rigorous evaluation of these new techniques, exploring different deep
architectures and comparing them on a common ground, identifying and disclosing 
important implementation details.
%
% This is carried out by: (1)
%fixing all other elements of the pipeline (the CNN and classifier learning
%framework and tuning), (2) disclosing all the implementation details, and (3)
%identifying those factors which are most significant in the performance of image
%classification methods, along with those aspects which are less critical. 
%
We identify several useful properties of CNN-based representations, including the
fact that the dimensionality of the CNN output layer can be reduced
significantly without having an adverse effect on performance. We also identify
aspects of deep and shallow methods that can be successfully shared.
In particular,
we show that the data augmentation techniques commonly applied to CNN-based methods
can also be applied to shallow methods, and result in an analogous performance boost.
Source code and models to reproduce the experiments in the paper is made publicly available.
%we are planning to provide the configurations and code that achieve the state-of-the-art
%performance on the PASCAL VOC Classification challenge, along with alternative configurations trading-off performance, computation speed and compactness.
%augmentation strategies on their performance, and
%comparing the impact that analogous strategies have when applied to a
%state-of-the-art feature encoding method, the Fisher
%Kernel~\cite{Perronnin10a}. This is carried out by: (1) fixing all other
%elements of the pipeline (CNN training framework, learning, tuning) (2)
%disclosing all the implementation details, and (3) identifying those factors
%which are most significant in the performance of a CNN-based image
%classification method, along with those aspects which are less critical, a
%consistent comparative analysis can be made, and the strengths of a CNN based
%approach to image classification when compared to shallow feature encoding
%approaches can be evaluated.
%
%Clearly describing each step of our methodology and taking a experimental
%approach, we achieve state-of-the-art performance over the PASCAL VOC
%dataset. Further, we demonstrate that the dimensionality of the CNN output layer
%can be reduced significantly without having adverse effect on performance. It is
%also shown that the data augmentation methods can also be applied to shallow
%feature encodings to achieve a boost in performance analogous to that observed
%with CNN-based methods, and in some ways the information captured by shallow and
%deep methods are complementary.

%% file: ieeetran_preamble.tex
\renewenvironment{figure}{\begin{figure*}}{%
    \end{figure*}\ignorespacesafterend% as suggested above
}
\renewenvironment{table}{\begin{table*}}{%
    \end{table*}\ignorespacesafterend% as suggested above
}

\newenvironment{subsectionpar}[1]{%
  \begingroup%
  \subsection{#1}}%
  {\endsubsection\endgroup}
  
\newenvironment{subsubsectionpar}[1]{%
  \begingroup%
  \subsubsection{#1}}%
  {\endsubsubsection\endgroup}
  
%\renewenvironment{paragraph}[1]{\subsubsection{#1}}{}

%-------------------------------------------------------------------------

\title{Return of the Devil in the Details:\\
Delving Deep into Convolutional Nets}

\author{Ken~Chatfield, Karen~Simonyan, Andrea~Vedaldi, and~Andrew~Zisserman\\
        Visual Geometry Group, Department of Engineering Science, University of Oxford \\        
        {\small \{ken,karen,vedaldi,az\}@robots.ox.ac.uk}
}
        
% KS: commented for the arxiv report; TODO: uncomment for PAMI submission
% \markboth{Return of the Devil in the Details}%
% {Chatfield \MakeLowercase{\textit{et al.}}: Delving Deep into Convolutional Nets}